\documentclass[letterpaper]{article} 
\usepackage{aaai23}  
\usepackage{times}  
\usepackage{helvet}  
\usepackage{courier}  
\usepackage[hyphens]{url}  
\usepackage{graphicx} 
\usepackage{natbib}  
\usepackage{caption} 
\usepackage{algorithm}
\usepackage{algorithmic}

\usepackage{newfloat}
\usepackage{listings}
\DeclareCaptionStyle{ruled}{labelfont=normalfont,labelsep=colon,strut=off} 
\floatstyle{ruled}
\newfloat{listing}{tb}{lst}{}
\floatname{listing}{Listing}
\title{fintech-kMC: Agent based simulations of financial platforms for design and testing of machine learning systems}
\author {
    Isaac Tamblyn,\textsuperscript{\rm 1,2,3}
    Tengkai Yu, \textsuperscript{\rm 1,4}
    Ian Benlolo \textsuperscript{\rm 1,3}
}
\affiliations {
    \textsuperscript{\rm 1} Cash App, Block, Toronto, ON M5A 1J7, Canada\\
    \textsuperscript{\rm 2} Vector Institute for Artificial Intelligence, Toronto, ON M5G 1M1, Canada \\
    \textsuperscript{\rm 3}  Department of Physics, University of Ottawa, Ottawa, ON K1N 6N5, Canada \\
    \textsuperscript{\rm 4}  Department of Computer Science, University of Victoria, Victoria, BC V8P 5C2, Canada \\

    itamblyn@squareup.com
}

\begin{document}

\maketitle

\begin{abstract}
We discuss our simulation tool, fintech-kMC, which is designed to generate synthetic data for machine learning model development and testing. fintech-kMC is an agent-based model driven by a kinetic Monte Carlo (a.k.a. continuous time Monte Carlo) engine which simulates the behaviour of customers using an online digital financial platform. The tool provides an interpretable, reproducible, and realistic way of generating synthetic data which can be used to validate and test AI/ML models and pipelines to be used in real-world customer-facing financial applications.
\end{abstract}

\emph{To appear at AAAI-23 Bridge Program: AI for Financial Services, Washington D.C., February 7 – 8, 2023}

\section{Introduction}

Simulated data is useful in machine learning (ML) work for developing and testing algorithms in a repeatable and verifiable way. We note that although historically there has been some debate about the utility of simulated data in machine learning applications given the challenges of sim2real~\cite{sim2real}, irrespective of whether simulated data is valuable for training \emph{production} models, they are extremely valuable in the initial stages of model design, unit testing, and verification processes. Because it is possible to have absolute certainty regarding ground truth labels as well as microscopic control over their data generation conditions, it is possible to remove the initial complexity and variability of noisy, real-world data, enabling a clean development environment to verify the functioning and logic of machine learning pipelines and models. Simulations are particularly valuable when the system under study is characterized by large class imbalances, rare events, and significant label noise. All of these challenging modelling problems frequently appear in fintech machine learning. 

This article is organized as follows. We briefly describe possible approaches to synthetic dataset generation. We then describe the kinetic Monte Carlo (kMC) algorithm in a general way. Next, we explain attributes that define agents in \texttt{fintech-kMC}, followed by a discussion related to the treatment of the rate constants which govern their behaviour. We provide examples of our simulation output, as well as some simple supervised machine learning tests which make use of this output. We discuss the limitations of our method, and finally, conclude. Our major contributions are:

\begin{itemize}
	\item a new agent-based simulation tool, driven by kinetic Monte Carlo is introduced
	\item a detailed description of the tool is provided, including its design, modelling capabilities, and output format
	\item demonstration of a predictive model trained to detect “bad actors” is given as a prototypical example of an ML model that uses our synthetic data
\end{itemize}

Different types of synthetic data generation protocols exist. They include simple heuristics (e.g. if an input feature takes some value $X$, set the output to label $Y$), simulations which can be deterministic ~\cite{example_deterministic_simulation} or stochastic~\cite{example_stochastic_simulation}, and generative models~\cite{gan, flow} which have been trained from sample data streams from the true system. We believe that generative models, while they may be able to best mimic a real-world dataset, introduce an entirely new set of issues since they themselves have dependencies on high-quality data pipelines, training infrastructure, monitoring, etc. A stochastic simulation is an excellent balance between being realistic enough to find implementation bugs in machine learning workflows yet simple enough not to introduce them.

\section{kinetic Monte Carlo (kMC)}

kMC has a long history of use in the physical sciences for describing the microscopic time evolution of a wide range of systems, such as molecules on catalytic surfaces~\cite{kMC_surface}, self-assembly and growth~\cite{kMC_self_assembly}, and radiation damage~\cite{kMC_radiation_damage}. The method has also been successfully applied to the social sciences, where it has been used to simulate the structure and information flow through large online social graphs such as Twitter~\cite{hashkat}.

\begin{algorithm}[h!]
\caption{kinetic Monte Carlo algorithm}
\label{alg:algorithm2}
\textbf{Parameters}: num\_agents, maximum\_time, etc\\
\textbf{Output}: logfile
\begin{algorithmic}[1] 
\WHILE{simulation\_time $<$ maximum\_time}
\STATE update\_rates()
\STATE rates := [$r_1, r_2, ..., r_n$]
\COMMENT{array of rates for $n$ events}
\STATE cummulative\_array = cumsum(rates/sum(rates))
\STATE $u_1$ := rand()
\COMMENT{Get random number $u_1\in(0,1]$}
\STATE event := binary\_search(cummulative\_array, $u_1$)
\STATE carry\_out\_event(event)
\STATE $u_2 := rand()$
\COMMENT{Get random number $u_2\in(0,1]$, update simulation time}
\STATE simulation\_time := simulation\_time $-ln(u_2) / R$
\STATE write out event information to logfile
\ENDWHILE
\end{algorithmic}
\end{algorithm}

Unlike some other Monte Carlo based simulation techniques (most notably Markov Chain Monte Carlo, MCMC), kMC is a rejection-free algorithm. An event occurs at every simulation step. Additionally, we note that the simulation step size, $\Delta t$, is not fixed. In a hypothetical system with only a single type of event, occurring at one characteristic rate, $r$, each step in simulation time will be approximately $\Delta t = r^{-1}$ apart. In a simulation with many possible events (and rates), the simulation will progress forward in time according to the total effective rate of the system (e.g. $R^{-1} = r_1^{-1} + r_2^{-1} + ...$). In practice, this means that as the total number of possible events grows, the time between each successive one decreases. This is exactly what happens in a real-world system. Consider an early-stage financial platform with exactly 1 customer. If that customer deposits money once per week, then after 1 month the transaction log will contain approximately 4 entries, spaced roughly 1 week apart. If the number of customers increases by x100 (and the timing of their deposities is randomly distributed) then after the same period of one month, we would expect 400 entries, with the average time between updates decreasing by x100. kMC simulations mimic this behaviour, and allow for irregularly spaced events and changing values of $\Delta t$ as the simulation progresses.

We note that in the above example, we assumed that the rates of client deposits were uncorrelated with each other and occurred at random times. In general this of course is not guaranteed to be true, particularly with payments which are associated with customer paycheques, rent, loan repayments, etc, which tend to occur around specific times of the month or the year. See Sec.~\textbf{Dynamic rates} for further discussion on how we account for this..

Disclaimer: in the following discussion, for pedagogical purposes, we give examples of actions or rules which might take place on a fintech platform. These are purely hypothetical, and are not intended to describe a recommended set of policies of a \emph{real} fintech platform. Such platforms are subject to a large number of policies and regulations (particularly with respect to ID requirements, customer age limits, etc). While \texttt{fintech-kMC} could be adapted to incorporate those specific requirements as needed, for this discussion we do not handle this complexity or region-specific limitations.

\section{Agents} 

Within \texttt{fintech-kMC}, customers are represented as agents. Currently, we have 2 distinct types of agents which are included within our model:

\begin{itemize}
	\item \emph{Individual customers}: These are the most common type of agent in typical simulations and represent individual customers. Within this type, several different archetypes exist (see Sec.~\textbf{Agent Archetypes}).

	\item \emph{Businesses}: Representing organizations or businesses, these agents engage in a restricted set of actions (e.g. they do not \texttt{pay\_rent}, \texttt{repay\_loan}, or \texttt{buy\_btc}). See Sec.~\textbf{Available actions} for further discussion of agent actions. Currently we have only a single archetype here.
\end{itemize}

Individual or business accounts can also be ``bad actors'' (e.g. engaged in credit card fraud, consumer scams, etc). Actions taken by these agents are primarily associated with positive labels. Subtle behavioural differences between such agents and regular, healthy, customers are what allow ML algorithms to make predictions.

\section{Static, Dynamical, and Triggered Rates}

\texttt{fintech-kMC} allows for the concurrent use of both static rates and dynamical rates. An example of a static rate could be the rate at which new customers join the platform. For the timescales we typically consider, this rate can be assumed to be constant. Conversely, dynamical rates can change over the course of the simulation. Some dynamical rates can change continuously, while others are activated according to a pre-established set of rules or when a particular simulation state has been reached (e.g. an agent successfully has their ID verified).

\textbf{Dynamic rates} Dynamical rate schedules can be based on time (e.g. the absolute simulated world-clock time, or a relative time, such as how many days a customer has been been a user of the platform). Regular events such as paycheque deposits are also included in this manner. For scheduled events such as paycheques, rent, etc, we set the rates of all affected agents to a very high value on days corresponding to such payments (e.g. 1$^{\textrm{st}}$ day of the month). This means the probability such an event will be selected is significantly higher than any other process. As soon as the event occurs, we reset this rate to 0. Using this approach, it is possible to naturally incorporate such ``guaranteed events'' into the kMC approach.

Other rules which govern dynamical rates can be based on the current state of the agent. For example, an agent may purchase bitcoin (BTC) at a frequency of $\approx 1/\textrm{week}$, \emph{provided their account balance is above a threshold} and they have \emph{passed the id verification process}. Similarly, an agent may make a monthly loan payment via a peer-to-peer (\texttt{p2p\_send}) payment \emph{until the balance is 0}. Allowing for dynamic and flexible modifications to the rate constants which determine agent behaviour allows for a realistic approach to modelling human behaviour.

\textbf{Available actions} Agents interact with one another and evolve through time via the actions they take within the simulation. Many such actions exist.

\begin{itemize}

\item \texttt{cash\_in}: money is moved into the account. The range of possible transfer amounts depends on whether or not the agent has completed \texttt{id\_verification}.

\item \texttt{cash\_out}: money is removed from the account. The range of possible transfer amounts depends on whether or not the agent has completed \texttt{id\_verification}.

\item \texttt{p2p\_send}: money is removed from the account and is transferred to another agent via a peer-to-peer payment.

\item \texttt{id\_verification}: successful verification of an agent ``unlocks'' other actions (specifically \texttt{btc\_buy}) and changes the maximum allowed \texttt{cash\_in} and \texttt{cash\_out} amounts. A customer will attempt to do this action until they are successful, with bad actors having a lower probability of success.

\item \texttt{btc\_buy}: after an agent has successfully verfied their ID via the \texttt{id\_verification} action, they can purchase BTC. This action reduces their cash balance and increases their BTC balance.

\item \texttt{pay\_rent}: if an agent pays rent, it will do so at a frequency of once every 30 days. 

\item \texttt{deposit\_paycheque}: if an agent receives a paycheque, it will be deposited into their account once every 14 days.

\item \texttt{repay\_loan}: loan repayments occur on a 7 day cycle (at a fixed percentage of the original amount) until the balance of the loan is zero.

\end{itemize}

We note that any actions involving money movements above can only occur if there is sufficient balance to do so (accounts cannot go negative).

Importantly, to represent the diversity of customer behaviour observed on fintech platforms by real customers, the individual rates each agent operates with are not identical, but rather are sampled randomly from a statistical distribution. We also use the concept of agent archetypes.

\subsubsection{Agent Archetypes}
While no two people are the same, in modelling and business analytics it is often useful to invoke the concept of customer archetypes. By understanding the behaviours and motivations of customers on our platform, we can segment them into groups which have similar actions and (likely) motivations. Within \texttt{fintech-kMC}, we make use of this idea by defining different agent archetypes which give rise to distinct behaviours.

For example, an agent archetype of a \texttt{cypto\_enthusiast} will be assigned a rate for BTC purchases from the normal distribution $N(\mu,\sigma)$, where $\mu = 2$ and $\sigma = 1$ (in units of $\textrm{days}^{-1}$), whereas a \texttt{crypto\_skeptic} would have that rate set to zero. The specific values of $\mu$ and $\sigma$ are adjustable input parameters that the modeller can use to best mimic the details of the particular fintech platform they wish to simulate. Similarly, a customer that is a \texttt{big\_spender} may have an average \texttt{cash\_in} amount which is significantly larger than their peers. Defining the agent archetype determines the distribution of rates and the typical range of money movement values an agent will use during the simulation. This allows for the creation of a meaningful diversity of agents within the dataset. Agents which belong to the same archetype will not have \emph{identical} behaviours, but rather their preferences and propensities will be sampled from the same statistical distributions.

\subsection{Output}

\texttt{fintech-kMC} provides simulation output in the form of structured files which should be familiar to practitioners working in the area of datascience and machine learning. They are \texttt{CSV} files consisting of timestamped records of all events and include information about the generating agent, the action that was taken, and any value associated with it. For example, when an agent has their ID verified, the following information is reported: time, initiating\_token, action, value. For example, this may appear as \texttt{2022-09-03 04:50:05.00, C\_83dhpqzz, id\_verification, True}. A p2p currency exchange includes an addtional field corresponding to the \texttt{receiving\_token}. For example: \texttt{2022-11-14 04:50:05.00, C\_b6589f, p2p\_sent, 2000, C\_83dhpqzz}.

Here \texttt{initiating\_token} is a hash of the agent’s unique numerical ID. Typically such tokens are used in real-world systems to uniquely specific customers without revealing any personally identifying information (PII) data. We use this formatting style to mimic the type of data one would find in raw log files from a real-world system as closely as possible. In a similar spirit, we note that a final dataframe has \texttt{NULL} values that must be accounted for (a common occurrence in realistic datasets) because different events can have different numbers of attributes. We intentionally provide the output data in this format where actions and agents are intermixed so that preprocessing steps used in the ML pipeline can also be validated and tested.

\section{Testing machine learning models}

\begin{figure}[h]
\centering
\includegraphics[width=0.9\columnwidth]{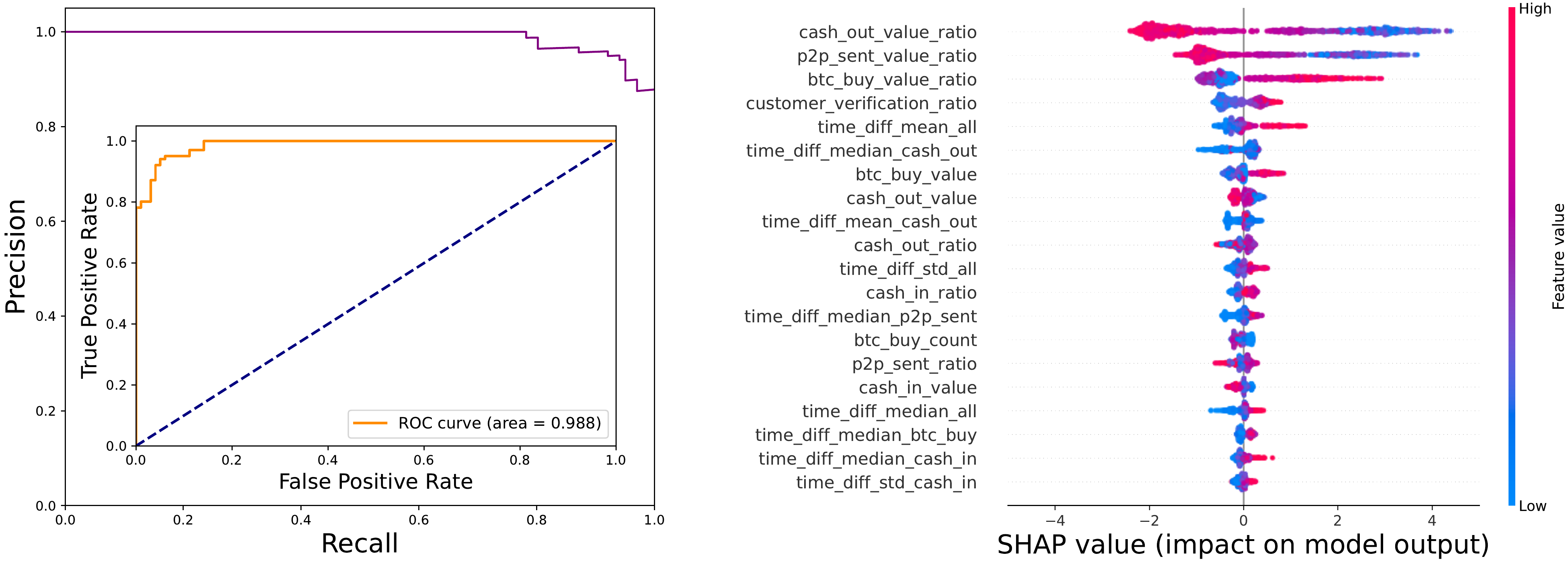} 
\caption{a) Precision-Recall and Receiver operating characteristic, ROC (inset) and b) SHAP~\cite{SHAP} analysis for our simple example binary classifier model. Model training and validation data were generated using the \texttt{fintech-kMC} tool. The full list of features used to train the model is listed in the online Supplementary Information.}
\label{f:xgboost}
\end{figure}

\begin{figure}[h]
\centering
\includegraphics[width=0.9\columnwidth]{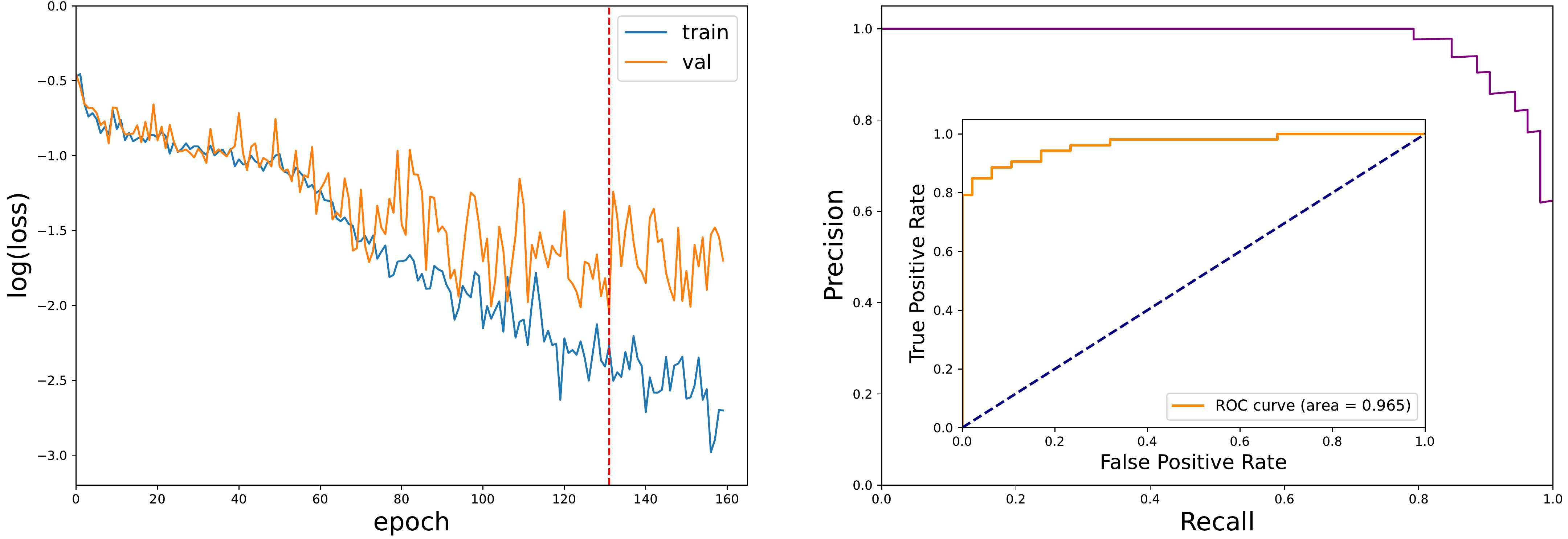} 
\caption{a) LSTM loss curves for train and validation sets b) Precision-Recall and ROC (inset) for the model (evaluated at epoch denoted by vertical red line). Hyperparameters are listed in the online Supplementary Information.}
\label{f:lstm}
\end{figure}

To demonstrate the utility of \texttt{fintech-kMC} with ML tools, here we show results from a toy experiment to detect and classify bad actors from regular customers. For this particular simulation, bad actors were configured to have a lower rate of ID verification success (50\% compared to 75\% for regular customers) and slightly different peer-to-peer money transfer behaviours (an average transfer amount of $5 \pm $3 vs $8 \pm $3 and a willingness to do such transfers only when their balance was above $15 \pm $3 vs $30 \pm $3). Here $\pm$ denotes the standard deviation of the normal distributions from which we sample to determine the specific transfer amounts. We created a population of 1000 agents where 50\% were bad actors and the remaining were regular, healthy customers. Our simulation covered a time period of $\approx 8$ days, which, given the various rate constants we used, represents approximately 100 actions for each agent. We use the convention that a positive label is deleterious (e.g. a scam, an account take-over, or some other activity we wish to identify and prevent). To train a model, we constructed a series of features listed in the online Supplementary Information. We used a train-validation-test split of 80\%-10\%-10\% and trained an XGBoost binary classifier~\cite{xgboost}, Fig.~\ref{f:xgboost}, as well as an LSTM~\cite{lstm}, Fig.~\ref{f:lstm}, on the data (full model training hyperparameters are provided in the online Supplementary Information).

\section{Limitations of the method}
\begin{itemize}
\item Introducing new types of actions requires more software implementation. We cannot learn directly from data streams from our real-world system.
\item Rate constant probability distributions in the control file are set manually by the user. To match them to a particular dataset, they must be optimized in a separate loop. Conceivably this could be achieved by automatic differentiation, but such capability is not currently included.
\item Large scale network effects are currently not well described due to the small number of agents we typically simulate using the tool (e.g. 1k). In principle, agent-based kMC can be used to simulate larger “worlds” (e.g. 1-10M) but for simplicity we have not implemented the typical optimization or parallelization techniques needed to achieve such performance here.
\item The current implementation is written in python. This is convenient for prototyping and adding new features, but results in a tool which is not as performant as it could be.
\item In the current implementation, actions can occur at any time during the simulated day, whereas in real-world data customers tend to exhibit daily trends. This is not, however, a fundamental limitation of the methodology.
\item Our positive labels are currently assigned at the customer level, rather than the individual transaction level. We are currently working to remove this restriction.
\item Data produced by the model should not be seen as a source of data for production models.
\end{itemize}

\section{Conclusions}
\texttt{fintech-kMC} is an agent-based model which can simulate the behaviour of customers of online digital financial platforms. The tool implements many actions typical of such platforms, such as peer-to-peer money movements, ID verification, and crypto purchases. By using kinetic Monte Carlo, events occur at realistic timescales and can have meaningful sequential dependencies. Data produced by the tool can be used to test and validate machine learning workflows in a controllable and repeatable way.

\bibliography{aaai23.bib}

\section{Supplementary Information}

Architexture details of our LSTM and XGBoost models are presented in Table~S\ref{t:xgboost} and Table~S\ref{t:lstm}. We also provide a list of the hand-designed features we used to train our feature-based ML models (Table~S\ref{t:xgboost_features}).

\begin{table}[!h]
\begin{tabular}{ll}
\hline
\texttt{base\_score}		&	0.5\\
\texttt{booster}			&	gbtree\\
\texttt{colsample\_bylevel}	&	1\\
\texttt{colsample\_bynode}	&	1\\
\texttt{colsample\_bytree}	&	1\\
\texttt{gamma}				&	0\\
\texttt{learning\_rate}		&	0.1\\
\texttt{max\_delta\_step}	&	0\\
\texttt{max\_depth}			&	3\\
\texttt{min\_child\_weight}	&	1\\
\texttt{missing}			&	None\\
\texttt{n\_estimators}		&	100\\
\texttt{nthread}			&	1\\
\texttt{objective}			&	binary$\colon$logistic\\
\texttt{reg\_alpha}			&	0\\
\texttt{reg\_lambda}		&	1\\
\texttt{scale\_pos\_weight}	&	1\\
\texttt{seed}				&	0\\
\texttt{subsample}			&	1\\
\texttt{verbosity}			&	1\\
\texttt{tree\_method}		&	hist\\
\hline
\end{tabular}
\caption{Parameters used to train our XGBoost~\cite{xgboost} model}
\label{t:xgboost}
\end{table}

\begin{table*}[!h]
\centering
\begin{tabular}{lllll}
Layer & Type               & Output units & Dropout rate & Return sequence \\ \hline
0     & Bidirectional LSTM & 64           & 0.2          & True            \\ \hline
1     & Bidirectional LSTM & 64           & 0.2          & True            \\ \hline
2     & Bidirectional LSTM & 64           & 0.2          & False           \\ \hline
3     & Dense              & 1            &              &                 \\ \hline
\end{tabular}
\caption{Architexture (layer definitions) of our LSTM~\cite{lstm} neural network model}
\label{t:lstm}
\end{table*}

\begin{table}[!t]
\begin{tabular}{l}
\hline
\texttt{total\_events} \\
\texttt{cash\_in\_count}	\\
\texttt{customer\_verification\_count} \\
\texttt{cash\_out\_count} \\
\texttt{p2p\_sent\_count} \\
\texttt{btc\_buy\_count} \\
\texttt{cash\_in\_ratio} \\
\texttt{customer\_verification\_ratio} \\
\texttt{cash\_out\_ratio} \\
\texttt{p2p\_sent\_ratio} \\
\texttt{btc\_buy\_ratio} \\
\texttt{cash\_in\_value} \\
\texttt{customer\_verification\_value} \\
\texttt{cash\_out\_value} \\
\texttt{p2p\_sent\_value} \\
\texttt{btc\_buy\_value} \\
\texttt{cash\_in\_value\_ratio} \\
\texttt{customer\_verification\_value\_ratio} \\
\texttt{cash\_out\_value\_ratio} \\
\texttt{p2p\_sent\_value\_ratio} \\
\texttt{btc\_buy\_value\_ratio} \\
\texttt{time\_diff\_mean\_all} \\
\texttt{time\_diff\_median\_all} \\
\texttt{time\_diff\_std\_all} \\
\texttt{time\_diff\_mean\_cash\_in} \\
\texttt{time\_diff\_median\_cash\_in} \\
\texttt{time\_diff\_std\_cash\_in} \\
\texttt{time\_diff\_mean\_customer\_verification} \\
\texttt{time\_diff\_median\_customer\_verification} \\
\texttt{time\_diff\_std\_customer\_verification} \\
\texttt{time\_diff\_mean\_cash\_out} \\
\texttt{time\_diff\_median\_cash\_out} \\
\texttt{time\_diff\_std\_cash\_out} \\
\texttt{time\_diff\_mean\_p2p\_sent} \\
\texttt{time\_diff\_median\_p2p\_sent} \\
\texttt{time\_diff\_std\_p2p\_sent} \\
\texttt{time\_diff\_mean\_btc\_buy} \\
\texttt{time\_diff\_median\_btc\_buy} \\
\texttt{time\_diff\_std\_btc\_buy} \\
\hline
\end{tabular}
\caption{Hand-designed features we created and used as input to our XGBoost model}
\label{t:xgboost_features}
\end{table}

\end{document}